\documentclass[letterpaper]{article} 
\usepackage{aaai24}  
\usepackage{times}  
\usepackage{helvet}  
\usepackage{courier}  
\usepackage[hyphens]{url}  
\usepackage{graphicx} 
\usepackage{natbib}  
\usepackage{caption} 
\usepackage{booktabs}
\usepackage{adjustbox}
\usepackage{threeparttable}
\usepackage{tablefootnote}
\usepackage{amsmath}
\usepackage{amssymb}
\usepackage{multirow}
\usepackage{algorithm}
\usepackage{algorithmic}
\usepackage{enumitem}
\usepackage{makecell}
\usepackage{stackengine}
\setenumerate[1]{itemsep=0pt,partopsep=0pt,parsep=\parskip,topsep=5pt}
\setitemize[1]{itemsep=1pt,partopsep=1pt,parsep=\parskip,topsep=4pt}

\newcommand{\best}[1]{\textbf{#1}}
\newcommand{\second}[1]{\underline{#1}}

\usepackage{newfloat}
\usepackage{listings}
\DeclareCaptionStyle{ruled}{labelfont=normalfont,labelsep=colon,strut=off} 
\floatstyle{ruled}
\newfloat{listing}{tb}{lst}{}
\floatname{listing}{Listing}
\title{Polar-Doc: One-Stage Document Dewarping with Multi-Scope Constraints under Polar Representation}
\author{
    Weiguang Zhang\textsuperscript{\rm 1},
    Qiufeng Wang\textsuperscript{\rm 1},
    Kaizhu Huang\textsuperscript{\rm 2},
}
\affiliations{
    \textsuperscript{\rm 1}Xi'an Jiaotong-liverpool University\\
    \textsuperscript{\rm 2}Duke Kunshan University\\
    
}

\usepackage{bibentry}

\begin{document}
\maketitle
\begin{abstract}
Document dewarping, aiming to eliminate geometric deformation in photographed documents to benefit text recognition, has made great progress in recent years but is still far from being solved. While Cartesian coordinates are typically leveraged by state-of-the-art approaches to learn a group of deformation control points, such representation is not efficient for dewarping model to learn the deformation information. 
In this work, we explore Polar coordinates representation for each point in document dewarping, namely \textbf{Polar-Doc}. In contrast to most current works adopting a two-stage pipeline typically, Polar representation enables a unified point regression framework for both segmentation and dewarping network in one single stage. Such unification makes the whole model more efficient to learn under an end-to-end optimization pipeline, and also obtains a compact representation. Furthermore, we propose a novel multi-scope Polar-Doc-IOU loss to constrain the relationship among control points as a grid-based regularization under the Polar representation. 
Visual comparisons and quantitative experiments on two benchmarks show that, with much fewer parameters than the other mainstream counterparts, our one-stage model with multi-scope constraints achieves new state-of-the-art performance on both pixel alignment metrics and OCR metrics. Source codes will be available at \url{*****}.
\end{abstract}

\section{1 Introduction}
\label{sec:intro}
Nowadays, people are more accustomed to digitizing paper documents (\textit{e.g.}, book pages, receipts, contracts, and leaflets) through smartphones rather than scanners. While such transition brings great convenience, it also leads to poor readability when the photographed document images suffer from serious geometric deformation (\textit{e.g.}, folded, curved or crumpled). To facilitate text recognition, it is desirable to eliminate such deformation by dewarping the deformed document into a flattened one. 
In addition, document dewarping is beneficial to many downstream tasks of document intelligence~\cite{cui2021document, yang2023modeling} such as document information extraction ~\cite{Chuwei2023GeoLayoutLM}, table structures parsing ~\cite{long2021parsing}, and document visual question answering ~\cite{mathew2021docvqa}. All of these lead to a growing demand for document dewarping.

\begin{figure}[t]
\centering
\includegraphics[width=0.48\textwidth] {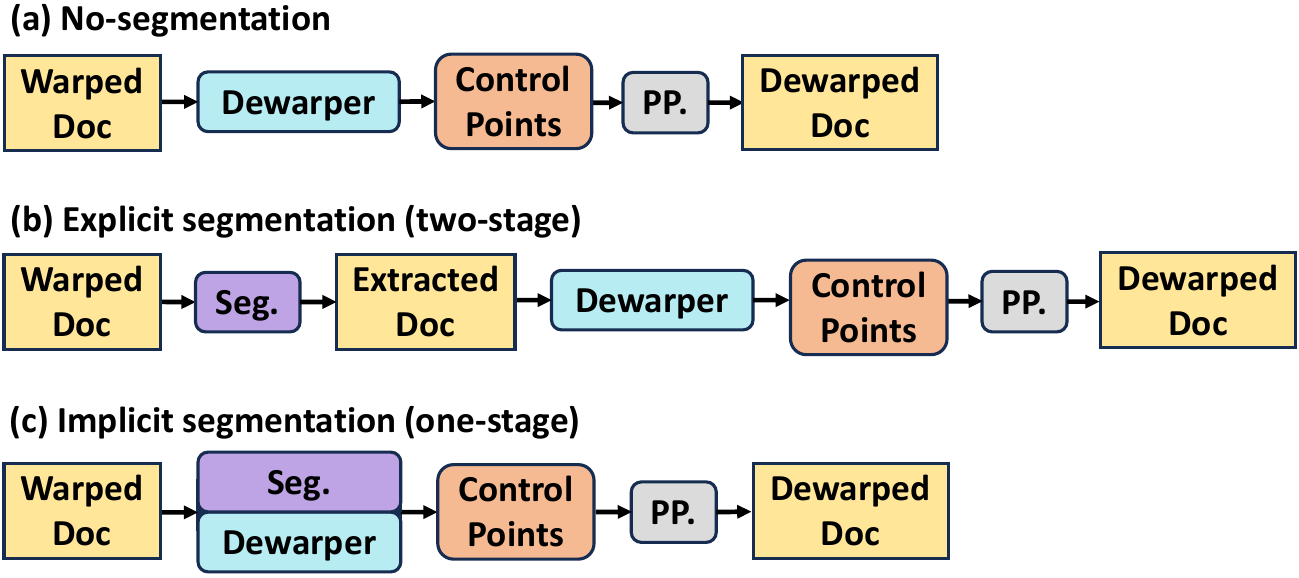}
\caption{Overview of different typical document dewarping pipelines. "Seg.", "Dewarper" and "PP." represent the document segmentation, document dewarping and post-processing block, respectively. (a) Simply dewarps the whole image without removing the background; (b) Explicitly designs a two-stage approach with document segmentation followed by a dewarper; (c) Our proposed Polar-Doc unifies the document segmentation and dewarping task into a one-stage control points prediction manner. 
}
\label{fig:first_example}
\end{figure}

As shown in Fig.~\ref{fig:first_example}, document dewarping task has been commonly regarded as a pixel-wise regression paradigm to predict mappings of control points coordinates between the warped and flatten document, including backward and forward mapping~\cite{ma2018docunet}. Essentially, this paradigm learns a group of 2D deformation control points to describe the geometric deformation of a warped paper. With such paradigm, document dewarping has achieved promising progress in recent years but is still far from being solved~\cite{das2019dewarpnet,xie2021document,feng2021doctr,jiaxin2022marior}. 
For example, the dewarped documents obtained by the recent state-of-the-arts are still distorted, as shown in Fig.~\ref{fig:compare_total}. 
In the following, we will examine the current document dewarping methods from three aspects to provide a clear understanding of the challenges in this task. 

Firstly, Cartesian coordinates are typically leveraged by the state-of-the-art document dewarping approaches to represent control points. Inspired by the success of Polar representation in instance segmentation \cite{RN77,RN76}, we argue that Polar coordinates are also more suitable for document dewarping thanks to the following three properties. (1) In the Polar coordinates, each point is represented by the angle and distance, where the angle can naturally capture document slantness. (2) It is flexible to represent points on the contour by the emitted rays from the Polar origin, making it efficient to represent the document contour (see Sec.~3.2). (3) It is efficient to approximately calculate the size of areas, then easily obtain the IOU (Intersection over Union) loss (see Sec.~3.3).  We also point out that though Cartesian representation may enjoy the first property by coordinates transformation, it lacks the merits due to the last two properties.

Secondly, early works~\cite{das2019dewarpnet,das2021end,ma2018docunet,li2019document,xie2021document} simply dewarp the whole image with both the background and document as shown in Fig.~\ref{fig:first_example}(a), leading to poor dewarping performance due to the background influence. To remove the background in the document dewarping, the state-of-the-arts methods~\cite{Feng2022geodocnet,feng2021doctr,jiaxin2022marior,RN58} usually rely on a two-stage approach, where a segmentation network is responsible for extracting the document boundary, then the extracted document is flattened by a dewarping network (see Fig.~\ref{fig:first_example}(b)). 
However, such two-stage approach is hard to optimize as they are trained separately. Moreover, any errors in the segmentation network might accumulate, resulting in dewarping mistakes. In this paper, we propose a one-stage method by integrating both segmentation and dewarping into a unified control points regression task as shown in Fig.~\ref{fig:first_example}(c). Specifically, we propose two parallel heads to jointly learn a group of control points including both contour points and mapping points as shown in Fig.~\ref{fig:pipeline}, where we transform the segmentation task into contour points regression task~\cite{RN77}.
We argue that our unified one-stage method has two benefits: (1) The supervision of both tasks could be end-to-end, making the model easier to optimize; (2) the backbone is shared and both heads are lightweight, resulting in a compact model with relatively fewer parameters.

Thirdly, it is inadequate to constrain the learned control points in current document dewarping methods. For instance, DocUNet~\cite{ma2018docunet} employs a global shift invariant loss to impose constraints between each point, resulting in a high computational cost with a quadratic complexity of $O(N^2)$ ($N$ indicates the number of control points). On the contrary, recent studies~\cite{xie2021document, jiang2022RDGR} have shifted their focus towards local shape constraints on neighboring points, which reduce computational costs significantly but may overlook global context. Since the predicted control points are spatially arranged in a grid format, such constraints are normally attributed to grid-based regularization~\cite{ma2018docunet,xie2021document,jiang2022RDGR}. 
Motivated by the IOU loss in object detection~\cite {yu2016unitbox,  zheng2020distance}, we propose a multi-scope Polar-Doc-IOU loss as our grid-based regularization that employs the Polar representation for efficient computation, where we consider both contour and patch-based losses to enforce both global and local alignment of control points simultaneously.

To sum up, our contributions are four-fold:
\begin{itemize}
  \item To the best of our knowledge, we are the first who explore Polar coordinates in document dewarping, making it flexibly to represent document contour and efficient to calculate IOU loss. 

  \item We propose a one-stage model to unify both segmentation and dewarping tasks in a joint regression framework of control points.

  \item We propose a Polar-Doc-IOU loss as a grid-based regularization term to constrain both global and local alignment of control points.
  \item Rich experiments show that state-of-the-art performance can be achieved with much fewer parameters in two challenging benchmarks.
\end{itemize}

\section{2 Related work}
\subsection{2.1 Traditional methods}
\label{sec:tridition}
Document dewarping has been widely investigated before the advent of deep learning. The most typical methods are usually 3D shape-based, which can be further divided into parametric and non-parametric models. Parametric models often rely on strong assumptions, such as general cylindrical surface (GCS)~\cite{RN13,RN17,RN18}, or developable surface~\cite{RN19,RN11}. Unfortunately, these strong assumptions limit the models to be used in special scenes. On the other hand, non-parameterized models abandon such assumptions and estimate accurate 3D representation, such as mesh~\cite{RN27,RN28,RN29}, point clouds~\cite{RN26}, and Non-Uniform Rational B-Splines~\cite{RN25}, while their models demand hardware heavily, like multi-view imaging~\cite{RN25,RN17,RN11}, laser range scanner~\cite{RN29} and structured light~\cite{RN21,RN26}, which fall behind the pace with the popularity of mobile camera.

\begin{figure*}[t]
	\centering
	\includegraphics[scale=0.69]{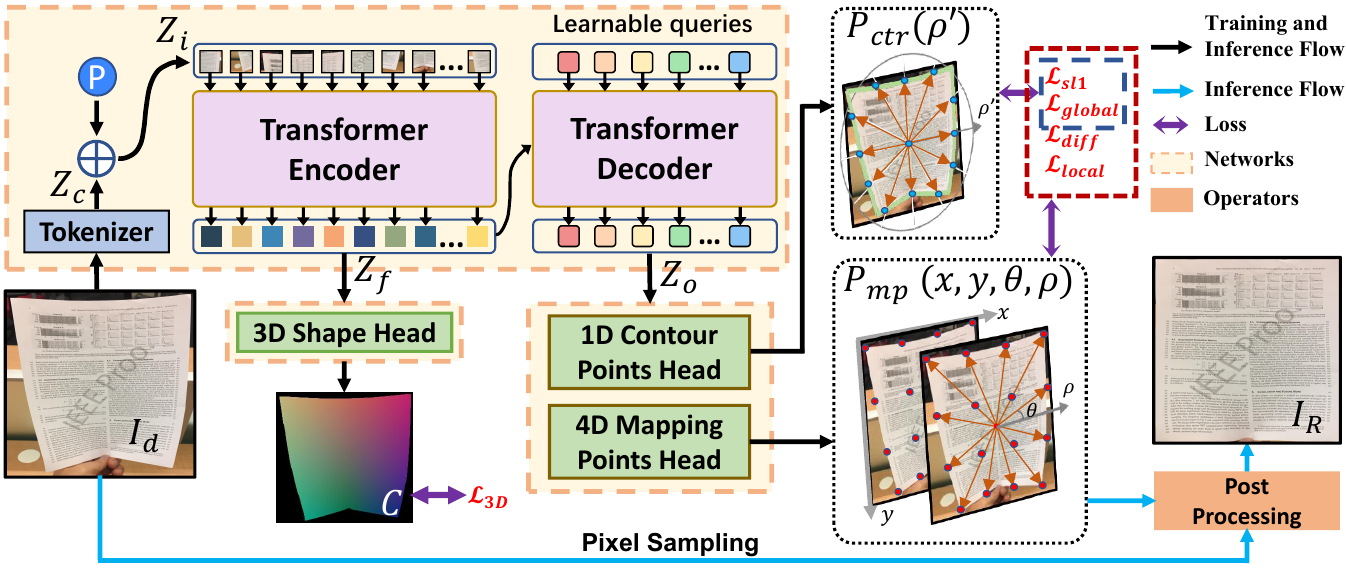}
	\caption{Overall pipeline of our proposed Polar-Doc. $I_{d}$, $C$, and $I_{R}$ denote the input warped document, predicted 3D shape coordinates, and final dewarped image respectively. $P_{ctr}$ and $P_{mp}$ are predicted contour points and mapping points respectively.}
	\label{fig:pipeline}
\end{figure*}


\subsection{2.2 Deep learning-based methods}
Recent success of deep learning in document dewarping contains great advances in data contribution, model design and regularization terms.

\textbf{Data contribution.} It is challenging to collect pairs of warped and flattened document images, leading to a scarcity of real-scene datasets for document dewarping. To address this issue, many early works, such as DIWF~\cite{xie2020dewarping}, DocUNet~\cite{ma2018docunet}, DDCP~\cite{xie2021document}, and DRIC~\cite{li2019document}, have focused on generating synthetic data with point-level annotation. Subsequently, Doc3D~\cite{das2019dewarpnet} and UVDoc~\cite{verhoeven2023neural} attempt to make synthetic documents more realistic  
by incorporating real shape information into the synthesizing process. More recently, studies have contributed pure-real datasets (e.g., WarpDoc~\cite{xue2022fourier} and DIW~\cite{RN58}), which however consist of page-level annotation only for weak supervision.

\textbf{Model design.}
We can divide current document dewarping models into two categories depending on whether or not to consider background removal by segmentation. For simplicity, early works(see Fig~\ref{fig:first_example}(a)) directly predict control points of the entire input image. For example, DewarpNet~\cite{das2019dewarpnet} and PDewarpNet~\cite{das2021end}, inspired by traditional 3D shape-based methods,  first predict 3D shape and then regress dense control points. Meanwhile, DocUNet~\cite{ma2018docunet}, DocProj~\cite{li2019document}, Inv3D~\cite{hertlein2023inv3d} and FDRNet~\cite{xue2022fourier} aim to improve the direct prediction of control points based on 2D information only. However, the redundant background should be excluded in practice. To this end, 
recent investigations (see Fig~\ref{fig:first_example}(b)) shift their focus towards two-stage models~\cite{Feng2022geodocnet,feng2021doctr, jiaxin2022marior,RN58,feng2023doctrp}. In the first stage, a semantic segmentation network is typically optimized to remove excess background; in the second stage, the extracted warped document image is fed into a dewarping network to predict control points~\cite{Feng2022geodocnet,feng2021doctr}. To improve dewarping performance, some studies~\cite{jiaxin2022marior,RN58} also integrate document boundary dewarping information in the second stage. Although such two-stage models meet practical requirements, they are hard to optimize end-to-end. In this paper, we propose a one-stage model that combines both segmentation and dewarping tasks into a unified control points regression framework, thus enabling a more efficient optimization. To the best of our knowledge, this is the first one-stage model that implicitly removes background for document dewarping.

\textbf{Grid-based regularization.}
To well learn control points in the document dewarping, researchers pay attention to grid-based regularization as the predicted control points are spatially arranged in a grid format. DocUNet~\cite{ma2018docunet} adopts a shift-invariant loss between any two points in the forward mapping, resulting in a huge computational cost. To tackle this issue,  DDCP~\cite{xie2021document} applies a Laplace operator on each control point relative to its neighboring points, which effectively guides the model to learn the local shape information. Recently, RDGR~\cite{jiang2022RDGR} summarizes the grid constraints and proposes a unified generalized energy minimization representation. Although these methods have demonstrated effectiveness, the constraints on the control points are not sufficient. Differently,  we propose in this paper a Polar-Doc-IOU loss to impose both global and local constraints on control points under Polar representation, thereby improving the learning of control points and achieving better dewarping performance.    


\subsection{2.3 Polar representation}
Polar representation has been extensively studied in segmentation tasks, such as StarDist~\cite{schmidt2018cell} for cell instance segmentation, DARNet~\cite{cheng2019darnet} for building segmentation, and  ESE-Seg~\cite{xu2019explicit} and Polar Mask++~\cite{RN77} for general segmentation tasks. In detail, Polar Mask++~\cite{RN77} divides the plane into equal angles; then, a general polygon can be represented as a series of learnable nodes with different Polar rays and fixed angles. Motivated by this, we learn a group of contour points to capture document contour in the Polar representation, and combine them with mapping points into a unified control points regression framework, forming a one-stage document dewarping model. 



\section{3 Methodology}

\subsection{3.1 Overall architecture}
As shown in Fig.~\ref{fig:pipeline}, our proposed Polar-Doc model consists of four components: (1) one encoder-decoder backbone for feature extraction, (2) a 3D shape head for auxiliary supervision, (3) two parallel heads to jointly learn contour points and mapping points for the segmentation and dewarping task respectively, (4) post-processing to obtain the dewarped document finally. In general, given a warped document image $I_d \in \mathbb{R}^{H\times W\times 3}$, our Polar-Doc predicts a series of sparse control points including contour points $P_{ctr}(\rho') \in \mathbb{R}^{a\times b\times 1}$ (i.e., there are $a$ layers with $b$ contour points in each layer) and mapping points $P_{mp}(x, y, \theta, \rho) \in \mathbb{R}^{h\times w\times 4}$. 
Furthermore, we adopt a post-processing block to obtain a pixel-by-pixel coordinate mapping $B \in \mathbb{R}^{H\times W\times 2}$ and final dewarped document image $I_R$. In the following, we will briefly describe each component. 

\textbf{Transformer encoder-decoder.}
Our backbone is based on the vision transformer (ViT)~\cite{Dosovitskiy2021vit} structure in the DocTr~\cite{feng2021doctr}. 
Specifically, we first employ a tokenizer to slice the document image into $8\times8$ patches, each of which is further fed into several residual convolution networks~\cite{he2016deep} to separately extract feature map $Z_c \in \mathbb{R}^{h\times w\times D}$, where $D$ is the dimension, and $h=\tfrac{H}{8}, w=\tfrac{W}{8}$. 
Next, we flatten $Z_c$ and add the cosine position encoding to obtain $Z_i \in \mathbb{R}^{N_t \times D}$, where $N_t$ represents the number of tokens. Token embedding obtained through six encoder layers and six decoder layers are denoted by $Z_f,Z_o \in \mathbb{R}^{N_t \times D}$ respectively.

\textbf{3D shape head.} 
3D shape can represent physical nature of paper deformation. Therefore, adding 3D information supervision can effectively improve the dewarping performance and has been widely applied~\cite{feng2021doctr, das2019dewarpnet}. 
Motivated by this, we also design an auxiliary head following the encoder output to estimate 3D coordinate information $C \in \mathbb{R}^{H\times W\times 3}$. Specifically, we first reshape encoder output $Z_{f}$. Then, we utilize bilinear interpolation and several convolutional layers to transform scale and dimension to match ground-truth 3D coordinate information $C^*$.


\textbf{Contour points and mapping points heads.}
In this paper, we predict two types of control points, including contour points and mapping points for document boundary and dewarping respectively.
By regarding contour points as a special type of control points, we can combine them with mapping points to form a unified points regression block.  
Concretely, we design two two-layer convolutional networks to predict 4D mapping points $P_{mp}(x, y, \theta, \rho) $ and 1D contour points $P_{ctr}(\rho')$ respectively, where $x,y$ represent Cartesian coordinates and $\theta$ represents the Polar angle for mapping points, $\rho,\rho'$ represent Polar radius for mapping points and contour points respectively. More details of Polar representation will be introduced in Sec.~3.2.

\textbf{Post-processing (PP.).} 
Based on mapping points $P_{mp}$ output from our Polar-Doc model, we first 
adopt Thin Plate Spline (TPS) interpolation~\cite{bookstein1989principal} to get a preliminary backward mapping $B' \in \mathbb{R}^{H'\times W'\times 2}$. Then, we apply bilinear interpolation to obtain the final backward mapping $B \in \mathbb{R}^{H_o\times W_o\times 2}$ with the same resolution of original image. Next, we utilize the \textit{grid\_sample} function in Pytorch to realize pixel-wise sampling and obtain the final dewarped document $I_R$. More details can be found in Appendix A.


\subsection{3.2 Polar-based one-stage model}
\label{sec:polar}
\begin{figure}[t]
	\centering
	\includegraphics[scale=0.085]{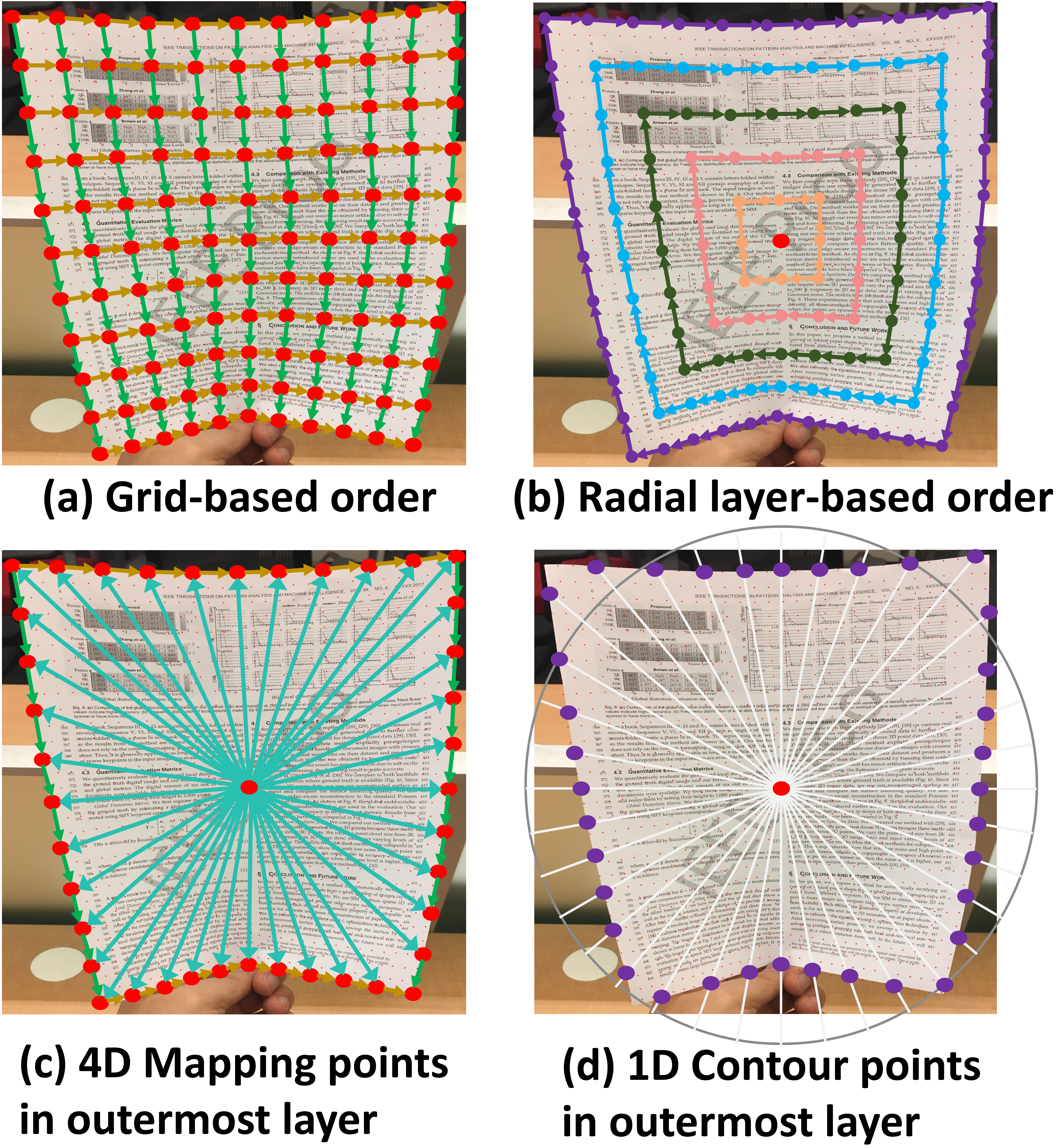}
	\caption{Different diagrams of points representation. (a) the standard rectangle grid order, (b) the radial arrangement, which constitutes a multi-layered outline form, (c) mapping points representation $P_{mp}(x, y, \theta, \rho)$, (d) contour points $P_{ctr}(\rho')$ with given equal angle interval sampling.}
	\label{fig:layer}
\end{figure}
\textbf{1D contour points representation}. 
In this paper, we propose an efficient way to implement semantic segmentation by contour points prediction in the Polar representation. 
Concretely, we represent multi-layer contour points as radial layer-based order shown in Fig.~\ref{fig:layer}(b), in contrast to the normal grid-based order in Fig.~\ref{fig:layer}(a). Under this multi-layer representation, the outermost contour can be used for semantic segmentation as shown in Fig.~\ref{fig:layer}(d). To efficiently represent contour points, we divide the whole plane into equal Polar angle interval $\theta$, then each contour point in the same layer can be simply represented by its Polar radius. In other words, we can omit angle information in the normal Polar coordinates for contour points due to the constant Polar angles. For example, the first point is angle $0$; the next one is angle $\theta$, $2\theta$, and so on. Finally, we can construct the 1D contour points representation $P_{ctr}(\rho')$. More details are in Appendix B.

\textbf{4D mapping points representation}.
Different from contour points, mapping points are mainly used for document dewarping. We extend the mapping points of common 2D Cartesian coordinates $P_{mp}(x, y)$ to 4D form $P_{mp}(x, y, \theta, \rho)$, and Fig.~\ref{fig:layer}(c) shows one example of the outermost layer. To be specific, we specify the center of mapping point as the Polar origin. Then, we calculate the distance from Polar origin to each mapping point as Polar radius, and the angle relative to the horizontal for each mapping point as Polar angle.

\textbf{Unified one-stage points regression framework}.
With two parallel heads of 1D contour points and 4D mapping points, both semantic segmentation and document dewarping are integrated into a one-stage regression task for control points, where the supervision on both contour points and mapping points is aggregated to optimize the backbone network. The detailed loss design will be further formulated in the next section.


\subsection{3.3 Polar-Doc-IOU loss}
\label{Sec_loss}
Although our Polar-Doc constitutes a unified one-stage regression model, it is non-trivial to accurately predict so many points. To this end, we propose a Polar-Doc-IOU loss as a new grid-based regularization term to improve the points regression. 
In the following, we first formulate the concept of Doc-IOU to overcome the limitation of box-based IOU in our framework, then we introduce some task-specific designs in Polar-Doc-IOU, including multi-scopes, edge regression, and focal mechanism.

In typical object detection areas, IOU loss functions~\cite{yu2016unitbox,rezatofighi2019generalized,zheng2020distance} that cooperate well with smooth-L1 loss~\cite{girshick2014rich}, have been widely used for the bounding box regression. 
However, these loss functions cannot  be applied straightforwardly in our task as our interested areas are usually irregular where the rectangle bounding box is not suitable and the area is not easy to calculate. To tackle this issue, we define our Polar IOU as the overlap of generalized sector area in the Polar representation:\scriptsize
\begin{equation}
	\label{eq:full_iou}
	{ \rm \operatorname{Polar-IOU}  } = \frac{\int_{0}^{2\pi} \frac{1}{2} \min(\rho,\rho^*)^2 d\theta}
	{\int_{0}^{2\pi}\frac{1}{2} \max(\rho,\rho^*)^2 d\theta},
\end{equation}\normalsize
where $\rho$ and $\rho^*$ are the predicted and ground-truth radius respectively with the Polar angle $\theta$ for each point within the same contour.
To simplify the computation, we discretize Eqn.~(\ref{eq:full_iou}) to obtain an approximate global Doc-IOU by following the work~\cite{RN77}:\scriptsize
\begin{equation}
	\label{eq:discrete_doc_iou}
	{  \rm \operatorname{Global\ Doc-IOU} } = \frac{\sum_{i=1}^{N}\rho^{\min}_i}
	{\sum_{i=1}^{N}\rho^{\max}_i}
	\approx \lim_{N \to \infty}\frac
	{\sum_{i=1}^{N}\frac{1}{2} \rho_{\min}^2 \Delta \theta_i}
	{\sum_{i=1}^{N}\frac{1}{2} \rho_{\max}^2 \Delta \theta_i}, 
\end{equation}\normalsize
where $N$ represents the number of points on the contour, $\rho^{\min}_i=\min(\rho_{i}^{},\ \rho_{i}^{*})$ and $ \rho^{\max}_i=\max(\rho_{i}^{},\ \rho_{i}^{*})$. 
By taking the negative log function, our global Polar-Doc-IoU loss can be represented as \scriptsize
\begin{equation}
	\label{eq:Doc_IoU_Loss}
	\mathcal{L}_{global\_d\_iou}
	= -\log \frac{\sum_{i=1}^{N}\rho^{\min}_i}
	{\sum_{i=1}^{N}\rho^{\max}_i}.
\end{equation}\normalsize

\begin{figure}[!t]
	\centering
	\includegraphics[scale=0.55]{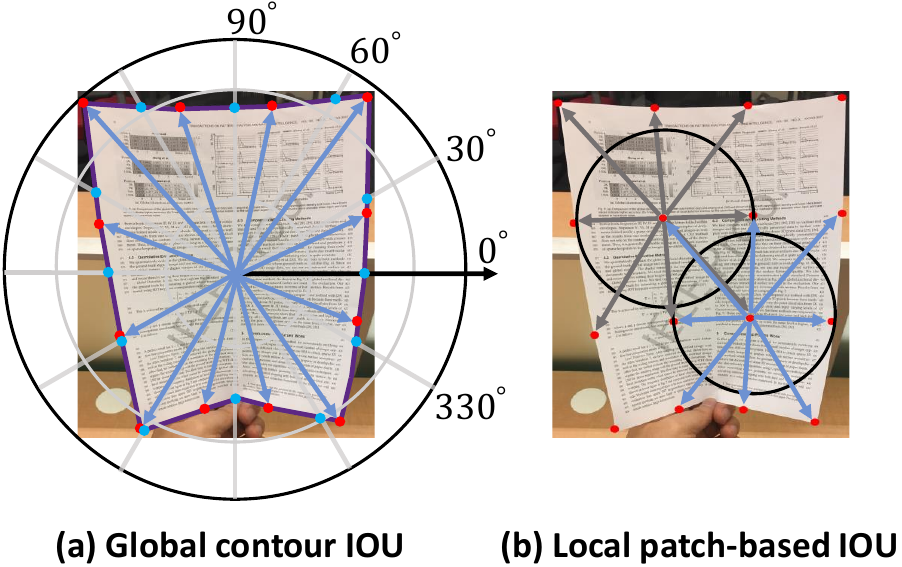}
	\caption{Polar-Doc-IOU diagram of both global and local scopes.
 (a) Mapping (red) and contour (blue) points diagram for the proposed global Polar-Doc-IOU loss. 
 (b) Two exemplary slideable $3\times3$ patches for the proposed local Polar-Doc-IOU loss. Each patch takes the local central point as the Polar origin and emits rays to represent a range of patch.}
	\label{fig:multi-scale}
\end{figure}

To be noted, we have multi-layer contour points. Therefore, our global Polar-Doc-IoU loss Eqn.~(\ref{eq:Doc_IoU_Loss}) can be applied to multi-scope constraints with different numbers of points $N$, and Fig.~\ref{fig:multi-scale} (a) shows an example of such loss on the outermost layer contour. In addition, we extend the global Doc-Polar-IoU loss to local $3\times3$ points patch, forming $\mathcal{L}_{local\_d\_iou}$ to consider the local scope constraints as illustrated in Fig.~\ref{fig:multi-scale} (b). The calculation of $\mathcal{L}_{local\_d\_iou}$ is the same as Eqn.~(\ref{eq:Doc_IoU_Loss}) except changing the number of points $N$ (e.g., $N=9$ for local scope).


Furthermore, motivated by the EIOU \cite{RN82} that regresses the length and height of the bounding box, we integrate the regression of each length of the four document edges into the global loss: 
\scriptsize{
\begin{align}
	\label{equ:edge}
	\mathcal{L}_{lrtb} =\left\|L-L^*\right\|_{1}+\left\|R-R^*\right\|_{1}+\left\|T-T^*\right\|_{1}+\left\|B-B^*\right\|_{1},
\end{align}}\normalsize
where $L,R,T,B$ represent the predicted left, right, top and bottom edge length of document respectively, and the symbol with "*" means the corresponding ground-truth length. Finally, the overall Polar-Doc-IOU loss could be expressed by:\scriptsize
\begin{align}
		\label{equ:global}
		\mathcal{L}_{{global}} &= (1 - \mathcal{L}_{global\_d\_iou})^{\gamma}(\mathcal{L}_{global\_d\_iou})+\mathcal{L}_{lrtb} \\
		\label{equ:local}
		\mathcal{L}_{local} &= (1 - \mathcal{L}_{local\_d\_iou})^{\gamma}(\mathcal{L}_{local\_d\_iou}).
\end{align}\normalsize
In the above, the factor $(1 - \mathcal{L}_{x})^{\gamma}$ is integrated to give more weight to hard-to-align points regression, which is motivated by the focal loss mechanism \cite{RN75}.


Derived from DDCP \cite{xie2021document}, we also introduce smooth L1 loss $\mathcal{L}_{sl1}$, differential coordinate loss $\mathcal{L}_{diff}$ and 3D shape regression loss $\mathcal{L}_{3D}$ formulated by:\scriptsize
\begin{align}\small
	  \label{eq:diff_loss}
		\mathcal{L}_{sl1} &= \frac{1}{N}\sum_{i}^{N} z_i,\  z_i = \begin{cases}0.5\left(p_i^{}-p_i^*\right)^2, \text { if }|p_i^{}-p_i^*|<1 \\ \left|p_i^{}-p_i^*\right|-0.5, \text { otherwise}\end{cases},\\
	  \mathcal{L}_{diff} &= \frac{1}{N}\sum_{i}^{N} 
  \left\|\delta_i^{}-\delta_i^*\right\|_{2},\  \delta_i = \sum_{j}^{k}(p_{j}^{}-p_{i}^{}),\\
	  \mathcal{L}_{3D} &= \left\|C - C^*\right\|_{1},
\end{align}\normalsize
where $k$ is the number of adjacent points, $p_i$ represents the $i$-th point, $p_{j}$ is the $j$-th adjacent point of the $p_i$, the term $p_{j}-p_{i}$ calculates the difference between two points, $C$ indicates predicted 3D coordinates, and the symbol with "*" means the corresponding ground-truth value. 

In summary, our final loss can be represented by\scriptsize
\begin{align}
	\label{equ:overall}
	&\mathcal{L} = \mathcal{L}_{sl1} + \alpha_1\mathcal{L}_{diff} +\alpha_2\mathcal{L}_{global} + \alpha_3\mathcal{L}_{local} + \alpha_4\mathcal{L}_{3D},
\end{align}\normalsize
where the last four terms form our new grid-based regularization. In the experiments, we will discuss the effectiveness of each component.

\section{4 Experiments}
\subsection{4.1 Datasets and experimental settings}
To evaluate the dewarping performance, we train our model on a widely used dataset Doc3D~\cite{das2019dewarpnet}, and test the model in two publicly available benchmark datasets including DocUNet~\cite{ma2018docunet} and DIR300 benchmark~\cite{Feng2022geodocnet}, which contain 130 and 300 warped and scanned image pairs, respectively. Compared to DocUNet, the DIR300 involves more complex background, various illumination conditions, and more challenging warped pattern. To be noted, we do not adopt the dataset WarpDoc~\cite{xue2022fourier} like most of its subsequent works (e.g., DocGeoNet~\cite{Feng2022geodocnet}, Marior~\cite{jiaxin2022marior}, PaperEdge~\cite{RN58}) as this dataset is mainly proposed for the document objects with scaling variation which is not our focus in this paper. 
 

During the training, we set the batch size to 6, and use Adam~\cite{kingma2014adam} as the optimizer along with the initial learning rate 1e-4. We conduct all experiments on 2 GTX1080Ti graphics cards with PyTorch 1.10.1. Various data augmentation methods are employed, including random cropping, background replacement, and color jitter. More details are in Appendix C.

\subsection{4.2 Evaluation metrics}
We adopt two widely used metrics to evaluate document dewarping models, including pixel alignment and OCR accuracy. For the pixel alignment, we mainly adopt the LD (Local Distortion) and one more robust metric AD (Aligned Distortion)~\cite{RN58}. For the reference, we also report the MS-SSIM metric although it is not suitable here due to its focus on capturing perceptual degradation (blur, noise, color shifts) rather than geometric distortions (feature point misalignment)~\cite{Feng2022geodocnet}.  
For the OCR accuracy, ED (Edit Distance) and CER (Character Error Rate) are widely used to quantify the similarity of two sequences. As there are not ground-truth string annotations, we adopt an OCR engine to recognize the ground-truth document image to obtain the reference string.

In the experiments, we find that the above-mentioned metrics are sensitive to the evaluation environments (e.g., Matlab)~\cite{ma2018docunet}. For fairness, we calculate all metrics under the same environment, i.e., MATLAB version 2022B and OCR engine Tesseract v5.1.0. Therefore, the reported numbers in Tab.~\ref{tab:comparisontable} and~\ref{tab:comparisontable2} may be different from those in the original papers due to different environments, but all results are based on their predicted documents directly from the github source of original papers, so it is fair.

\begin{figure*}[!t]
	\centering
	\includegraphics[width=0.9\textwidth]{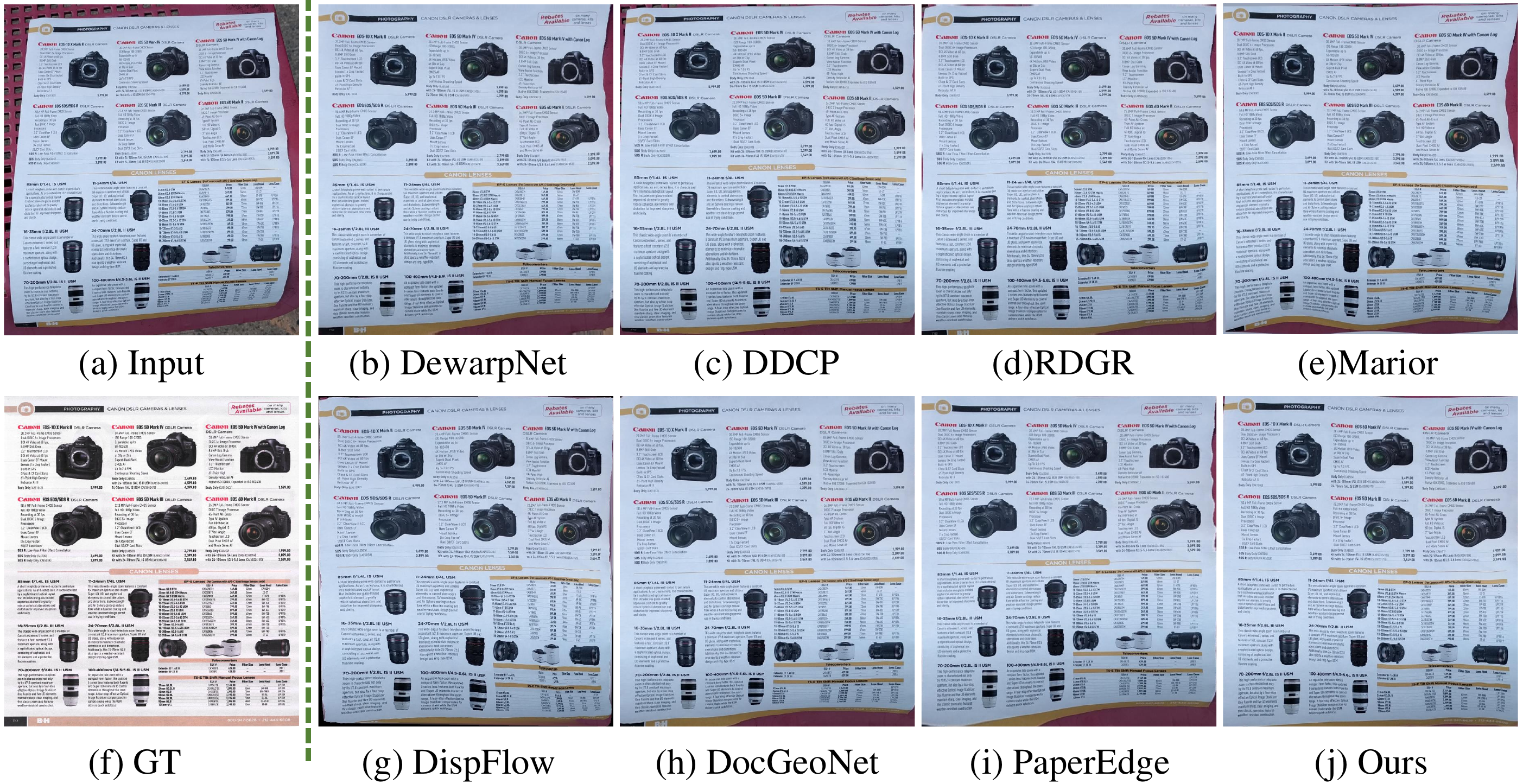}
 \caption{Visual qualitative comparisons. (a) Input,  (b) DewarpNet~\cite{das2019dewarpnet}, (c) DDCP~\cite{xie2021document}, (d) RDGR~\cite{jiang2022RDGR},  (e) Marior~\cite{jiaxin2022marior}, (f) Ground-Truth, (g) DispFlow~\cite{xie2020dewarping}, (h) DocGeoNet~\cite{Feng2022geodocnet} ,(h) PaperEdge\cite{RN58} and (j) our Polar-Doc. More samples are in Appendix D.}
	\label{fig:compare_total}
\end{figure*}

\subsection{4.3 Comparison with other SOTA methods}
In this subsection, we compare the performance of proposed Polar-Doc model with recent state-of-the-art (SOTA) works on two benchmarks, both qualitatively and quantitatively. 


\begin{table}[ht]
	\centering
	\caption{Quantitative dewarping performance comparisons on the \hbox{DocUNet} benchmark dataset. \textbf{Bold} indicates the best, \underline{underline} indicates second-best. The last column shows the network size by the number of parameters (millions).}
	\resizebox{\linewidth}{!}{%
		\begin{tabular}{l c c c c c c c c c}
			\toprule
			Method &MS-SSIM $\uparrow$ & LD $\downarrow$ & AD $\downarrow$ &  CER $\downarrow$ & ED $\downarrow$ & Para. $\downarrow$ \\
			\midrule
			Deformed  &0.246 & 20.51 & 1.026 & 0.481 & 1756.86 & - \\
			DewarpNet~\cite{das2019dewarpnet} &0.472 & 8.41 & 0.412 & 0.286 & 1012.34 & 86.9M\\ 
			DispFlow~\cite{xie2020dewarping}&0.431 & \second{7.64} & 0.411 & 0.356 &  1415.5 & 23.6M  \\ 
			DocTr~\cite{feng2021doctr}&\best{0.511} & 7.77 &  0.404 & 0.278 & 995.78 & 26.9M \\ 
			DDCP~\cite{xie2021document} &0.474 & 8.92 & 0.459  & 0.281 & 1048.50 & \second{13.3M} \\ 
			RDGR~\cite{jiang2022RDGR}&0.496 & 8.53 & 0.453 & 0.274 & 1003.06 & - \\
			Marior~\cite{jiaxin2022marior}&0.448  & 8.42 & 0.470 & 0.283 & 999.54 & - \\
			PaperEdge~\cite{RN58} &0.472 & 8.01 & \second{0.385} & \second{0.253} & \second{910.18} & 36.6M  \\ 
			DocGeoNet~\cite{Feng2022geodocnet} &\second{0.504 } & 7.70 & 0.395 & 0.279 & 990.76 & 24.8M \\ 
            Inv3d~\cite{hertlein2023inv3d} &0.443 & 9.32 & 0.512 & 0.312 & 1138.26 & 25.7M  \\\bottomrule 
			Ours &0.499 & \best{7.24} & \best{0.382} & \best{0.246} & \best{868.88} & \best{9.6M} \\
	\end{tabular}}
	\label{tab:comparisontable}
\end{table}

\begin{table}[t]
	\centering
	\caption{Quantitative dewarping performance comparisons on the \hbox{DIR300} benchmark dataset.}
	\resizebox{\linewidth}{!}{
		\begin{tabular}{l c c c c c c c c c}
			\toprule
			Method & MS-SSIM$\uparrow$ & LD $\downarrow$ & AD $\downarrow$ &  CER $\downarrow$ & ED $\downarrow$ & Para. \\
			\midrule
			Deformed  & 0.317 & 39.58 & 0.771 & 0.372 & 1434.54 & - \\
			DewarpNet~\cite{das2019dewarpnet} & 0.492 & 13.94 & 0.332 & 0.281 & 1058.36 & 86.9M \\ 
			DocTr~\cite{feng2021doctr}& \second{0.620} & 7.23 &  0.256 & \second{0.173} & \second{589.98} & 26.9M \\ 
			DDCP~\cite{xie2021document} & 0.552 & 10.97 & 0.356  & 0.554 & 2189.93 & \second{13.3M} \\ 
			DocGeoNet~\cite{Feng2022geodocnet} & \best{0.628} & \best{6.28} & \second{0.219} & 0.229 & 836.31 & 24.8M \\ \bottomrule 
			Ours & 0.605 & \second{7.17} & \best{0.206} & \best{0.170} & \best{549.64} & \best{9.6M}
	\end{tabular}}
	\label{tab:comparisontable2}
\end{table}

\textbf{Quantitative comparisons.} 
The quantitative results on DocUNet benchmark~\cite{ma2018docunet} are shown in Tab.~\ref{tab:comparisontable}. The first row "Deformed" means that the warped documents are directly input for the evaluation without using any dewarping methods, hence its metrics are the worst. From Tab.~\ref{tab:comparisontable}, we can see that our proposed model obtains the best performance on all evaluation metrics except the MS-SSIM, which is not a suitable metric here~\cite{Feng2022geodocnet}. Specifically, LD is decreased largely from 7.64 to 7.24. In addition, the size of our model is the smallest thanks to the proposed unified one-stage regression model, which is crucially important for the practical application.  
Meanwhile, we also compare the results on the most recent benchmark DIR300 in Tab.~\ref{tab:comparisontable2}, and we can get the same conclusion. Due to the open-source issue of several models in Tab.~\ref{tab:comparisontable}, we do not compare them in Tab.~\ref{tab:comparisontable2}. In summary, the quantitative comparisons verify that our Polar-Doc model can achieve state-of-the-art performance with much fewer parameters.


\textbf{Visual qualitative comparisons.} 
Fig.~\ref{fig:compare_total} shows one example to compare the dewarping performance qualitatively, and more samples of various types of paper documents (e.g., receipts, leaflets and books) are shown in Appendix D. We can see that our proposed Polar-Doc model demonstrates superior dewarping quality, especially in local content and complete edge extraction.


\subsection{4.4 Ablation studies}
In this section, we conduct the ablation studies on the DocUNet benchmark from three aspects. 

In our Polar-Doc, we use 4D form $P_{mp}(x, y, \theta, \rho)$ to represent mapping points. To verify the effectiveness of the Polar representation, we conduct one experiment by simply removing Polar coordinates to form a 2D Cartesian representation for mapping points. Due to unavailable Polar representation, 1D contour points head is not considered in the 2D form, resulting in the global Polar-Doc-IOU loss $\mathcal{L}_{global}$ is removed in Eqn.~(\ref{equ:overall}). It is shown in Tab.~\ref{tab:ablation1} that 4D form representation improves performance in all metrics. In addition, we show some examples with relatively large slopes to qualitatively demonstrate the excellent representation ability of Polar coordinates for document slantness, and the detailed visualization can be found in Appendix E. 
\scriptsize
\begin{table}[t]
	\centering\caption{Ablation study for two types of representations: $\boldsymbol{2D}$- using Cartesian representation only in the mapping points head; $\boldsymbol{4D}$- integrating Polar representation.}
	\resizebox{0.7\linewidth}{!}{ 
		\begin{tabular}{cc|cccc}
			\hline
			$\mathrm{2D}$ & $\mathrm{4D}$ & $\mathrm{LD} \downarrow$ & $\mathrm{AD} \downarrow$ & $\mathrm{CER} \downarrow$ & $\mathrm{ED}$ $\downarrow$ \\
			\hline $\checkmark$ & & 8.59 & 0.438 & 0.277 & 999.56 \\
			& $\checkmark$ & $\mathbf{7.24}$ & $\mathbf{0.382}$ & $\mathbf{0.246}$ & $\mathbf{868.88}$\\
			\hline
	\end{tabular}}
	\label{tab:ablation1}
\end{table}\normalsize

\begin{table}[!ht]
	\centering
	\caption{Comparison of different grid-based regularization terms on the DocUNet benchmark with the same backbone.}
    \resizebox{\linewidth}{!}{
	\begin{tabular}{lcccc}
		\toprule
		\textbf{Method} &  \textbf{LD} $\downarrow$ & \textbf{AD} $\downarrow$ &\textbf{CER} $\downarrow$ & \textbf{ED} $\downarrow$\\
		\midrule
		Shift Invariant loss~\cite{ma2018docunet} & 15.48 & 0.843 &0.376 & 1501.32 	\\
		Differential Coordinates~\cite{xie2021document} & 8.76	& 0.440 & 0.281  & 	1002.48 \\
		 Generalized Energy Minimization~\cite{jiang2022RDGR} & 8.71 & 0.443 & 0.279 & 1008.98  \\
		Ours & \textbf{7.24} & \textbf{0.382} & \textbf{0.246} & \textbf{868.88}  \\
  \hline
	\end{tabular}}
	\label{tab:bound}
\end{table}

\begin{table}[!h]
	\scriptsize
	\centering\caption{Ablation study for different Polar-Doc components: $\boldsymbol{4D}$ - baseline model with only 4D mapping points head; $\boldsymbol{3D}$ - 3D shape head; $\boldsymbol{M_{g}}$ - incomplete global Polar-Doc-IOU loss $\mathcal{L}_{global\_d\_iou'}$ with mapping points head; $\boldsymbol{M_{l}}$ - local Polar-Doc-IOU loss $\mathcal{L}_{local\_d\_iou}$; $\boldsymbol{FM}$ - focal mechanism; $\boldsymbol{E}$ - document edge regression term $\mathcal{L}_{lrtb}$; $\boldsymbol{C_{g}}$ - add contour points head for global Polar-Doc-IOU loss $\mathcal{L}_{global}$.}
	\resizebox{\linewidth}{!}{
		\begin{tabular}{|c|c|c|c|c|c|c|c|c|c|c|}
			\hline \multicolumn{7}{|c|}{ Polar-Doc Components } & \multicolumn{4}{c|}{ Experimental Results } \\
			\hline $\boldsymbol{4D}$ & $\boldsymbol{3D}$ & $\boldsymbol{M_{g}}$ & $\boldsymbol{M_{l}}$ & $\boldsymbol{FM} $ & $\boldsymbol{E}$ & $\boldsymbol{C_{g}}$ & LD$\downarrow$ & AD $\downarrow$ & CER $\downarrow$ & ED $\downarrow$\\
			\hline$\checkmark$ & & & & & & & $9.43$ & $0.490$ & $0.338$ & $1410.06$\\
			\hline$\checkmark$ & $\checkmark$ & & & & & &$8.67$ & $0.483$ & $0.326$ & $1363.20$ \\
			\hline$\checkmark$ & $\checkmark$ & & $\checkmark$ & & & &$7.89$ & $0.415$  & $0.283$ & $1020.44$\\
			\hline$\checkmark$ & $\checkmark$ & $\checkmark$ & & & & &$7.50$ & $0.421$   & $0.267$ & $975.32$\\
			\hline$\checkmark$ & $\checkmark$ & $\checkmark$ & $\checkmark$ & & & &$7.46$ & $0.402$ & $0.263$ & $947.84$\\
			\hline$\checkmark$ & $\checkmark$ & $\checkmark$ & $\checkmark$ & $\checkmark$ & & &$7.42$ & $0.395$ & $0.260$ & $940.90$\\
			\hline $\checkmark$ & $\checkmark$ & $\checkmark$ & $\checkmark$ & $\checkmark$ & $\checkmark$ & & $7.41$ & $0.393$ & $0.259$ & $941.42$\\
			\hline $\checkmark$ & $\checkmark$ & $\checkmark$ & $\checkmark$ & $\checkmark$ & $\checkmark$ & $\checkmark$ & $7.24$ & $0.382$ & $0.246$ & $868.88$\\
			\hline
	\end{tabular}}
	\label{tab:ablation2}
\end{table}

To show the effectiveness of our proposed Polar-Doc-IOU loss in Eqn.~(\ref{equ:overall}), we compare it with other grid-based regularization terms with the same backbone model in Tab.~\ref{tab:bound}. Compared to the previous methods, our Polar-Doc-IOU loss can extensively consider both local and global constraints, thus achieving the best performance on both pixel alignment and OCR accuracy metrics. The detailed analysis are in Appendix E. In the following, we will verify the effectiveness of each component in our proposed Polar-Doc.

As demonstrated in Tab.~\ref{tab:ablation2}, the first row shows that vanilla 4D mapping points head without any extensive designs can work for document dewarping, while it performs poorly. 
In the second row, 3D shape regression head produces a certain improvement, \textit{e.g.}, decreasing LD from 9.43 to 8.67. 
Row 3-5 compare local and global constraints in the Polar-Doc-IOU loss. Both of them achieve apparent performance improvement, but $\mathcal{L}_{local\_d\_iou}$ performs relatively better on AD, while $\mathcal{L}_{global\_d\_iou'}$ better on LD conversely. We attribute this phenomenon to the difference in weight of SIFT flow calculated by LD and AD. 
Combining both, we can obtain an even better effect in the fifth row.
We further introduce focal mechanism in the sixth row and add edge regression in the seventh row,  both attaining slightly higher performance. Last, adding 1D contour points head leads to a further improvement, especially on LD. In summary, we can confirm that each component is effective and complementary to help achieve new state-of-the-art performance.




\section{5 Conclusion}
In this paper, we propose a simple yet effective method, Polar-Doc, to dewarp document images in a one-stage manner. To the best of our knowledge, this is the first attempt to explore Polar coordinates in document dewarping, where we can efficiently combine document segmentation and dewarping tasks into one unified control points regression framework, forming the first one-stage model. Furthermore, we propose a novel Polar-Doc-IOU loss to constrain both global and local alignment of control points. 
Extensive experimental results show that our proposed Polar-Doc model can achieve  new state-of-the-art performance on both pixel alignment and OCR accuracy even with much fewer parameters.
In the future, we will explore content-based information to better empower photographed documents recognition.

\bibliography{aaai24}
\end{document}